\documentclass{article}
\usepackage{ijcai20}

\usepackage{times}
\usepackage{soul}
\usepackage{url}
\usepackage[hidelinks]{hyperref}
\usepackage[utf8]{inputenc}
\usepackage[small]{caption}
\usepackage{graphicx}
\usepackage{amsmath}
\usepackage{booktabs}
\usepackage{algorithm}
\usepackage{algorithmic}

\usepackage{xcolor}
\usepackage{courier}
\usepackage{amsmath}
\usepackage{amssymb}
\usepackage{graphicx}
\usepackage{subfigure}
\usepackage{longtable}
\usepackage{rotating}
\usepackage{multirow}
\usepackage{multicol}
\usepackage{xcolor}
\usepackage{fancyhdr}
\usepackage{enumerate}
\usepackage{float}
\usepackage{url}
\usepackage{balance}
\usepackage{makecell}
  \newcommand\figcaption{\def\@captype{figure}\caption}
  \newcommand\tabcaption{\def\@captype{table}\caption}
\makeatother
\title{Learning Robust Data Representation: A Knowledge Flow Perspective}

\author{
Zhengming Ding$^\dagger$\and
Ming Shao$^\ddagger$\and
Handong Zhao$^\sharp$ \and
Sheng Li$^\natural$ 
\affiliations
$^\dagger$Department of CIT, Indiana University–Purdue University Indianapolis, IN\\
$^\ddagger$Department of Computer and Information Science, University of Massachusetts Dartmouth, MA\\
$^\sharp$Adobe Research, San Jose, CA\\
$^\natural$Department of Computer Science, University of Georgia, GA\\
\emails
{zd2@iu.edu, mshao@umassd.edu, hazhao@adobe.com, sheng.li@uga.edu}
}

\begin{document}

\maketitle


\begin{abstract}
It is always demanding to learn robust visual representation for various learning problems; however, this learning and maintenance process usually suffers from noise, incompleteness or knowledge domain mismatch. Thus, robust representation learning by removing noisy features or samples, complementing incomplete data, and mitigating the distribution difference becomes the key. Along this line of research, low-rank modeling has been widely-applied to solving representation learning challenges. This survey covers the topic from a knowledge flow perspective in terms of: (1) robust knowledge recovery, (2) robust knowledge transfer, and (3) robust knowledge fusion, centered around several major applications. First of all, we deliver a unified formulation for robust knowledge discovery given single dataset. Second, we discuss robust knowledge transfer and fusion given multiple datasets with different knowledge flows, followed by practical challenges, model variations, and remarks. Finally, we highlight future research of robust knowledge discovery for incomplete, unbalance, large-scale data analysis. This would benefit AI community from literature review to future direction.
\end{abstract}

\section{Introduction}

\begin{figure}[t]
\begin{center}
      \hspace{-1mm}\includegraphics[width=0.48\textwidth]{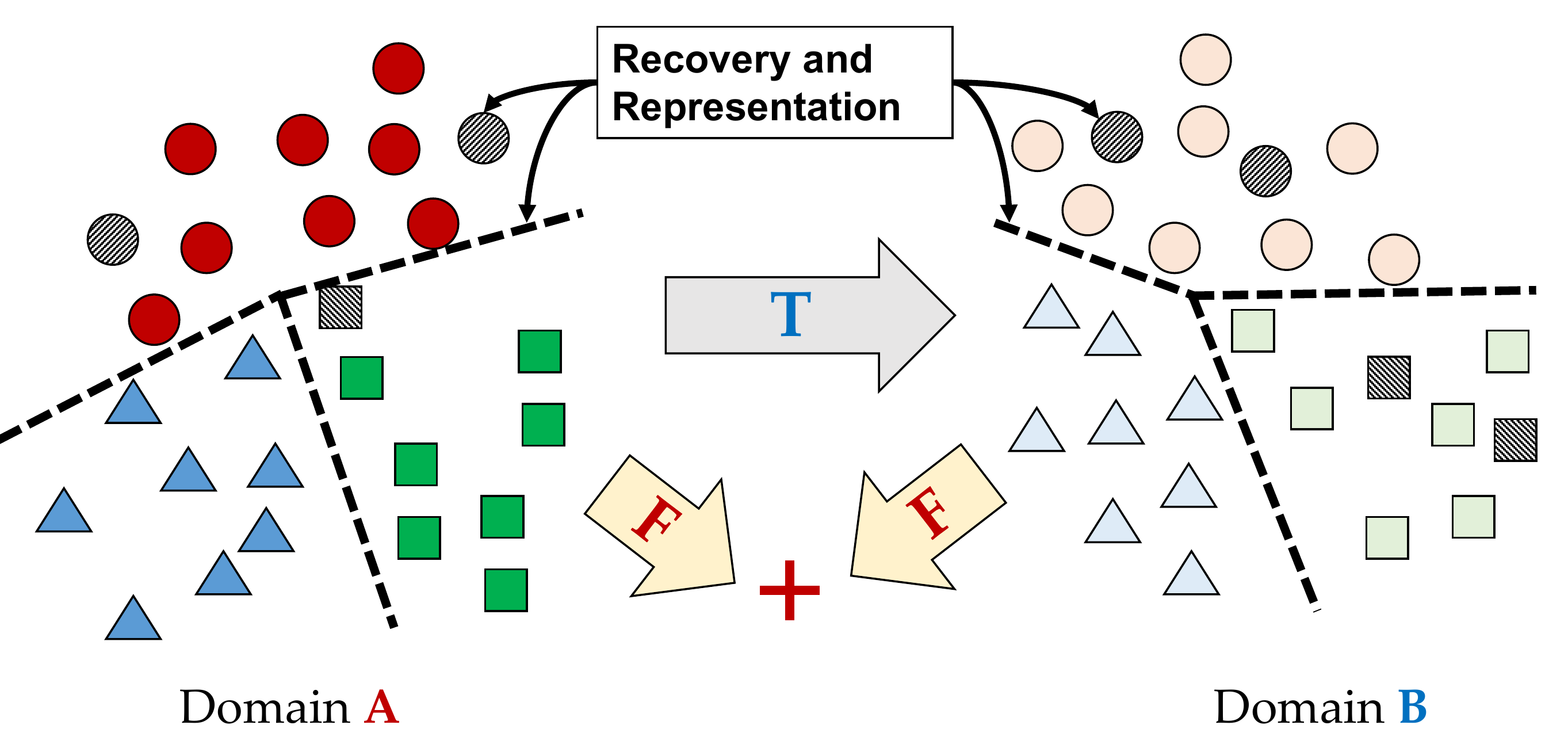}\vspace{-2mm}
  \caption{Illustration of our robust representation learning from a knowledge flow perspective in three fashions including knowledge recovery for single domain/view (dash line for discriminant representation and grey samples for outliers and noises), knowledge transfer for multiple views/domains (arrow with $\mathbf{T}$) and knowledge fusion for multiple views/domains (arrows with $\mathbf{F}$).}\label{fig1}\vspace{-5mm}
\end{center}
\end{figure}

In various learning tasks, knowledge recovery from given datasets usually refers to identifying discriminative features, or labels through supervised or unsupervised approaches. The data itself may come from either a single knowledge domain or multiple domains. In addition, data would suffer from several issues, e.g., noises, outliers, missing data, and domain divergences for multi-domain data analysis. How to deal with those challenges while still be able to discover meaningful knowledge is the key issue. It is well-known that low-rank modeling is one such robust knowledge discovery tool widely-used in the variety of vision, data mining, and machine learning problems. Basically, based on the number of knowledge domains and knowledge flow, we reshape robust representation learning from a knowledge flow perspective into three groups: (1) robust knowledge recovery for single domain, (2) robust knowledge transfer and (3) robust knowledge fusion for multiple domains. Their meanings and connections are illustrated in Figure \ref{fig1}.

First, robust knowledge recovery aims to recover the knowledge from single-domain data, assuming the data is generated from some unknown support(s) while suffered from corruption and noises. Such an assumption allows us to recover the knowledge by decomposing the data into low-rank and sparse components \cite{candes2011robust,liu2013robust}, which helps many high demanding applications such as data clustering, outlier detection, image segmentation, and classification. The low-rank modeling is usually integrated with embedding, representation learning \cite{li2016learning,ding2016robustaaai,ding2016deep}, and is able to work with vector data or graph \cite{shao2016scalable}.

Second, robust knowledge transfer targets at adapting knowledge from a well-labeled external domain to a target domain through coupling the distribution mismatch. The key is to align different domains through low-rank modeling and thus smoothly transit knowledge to the target domain, regardless of noises, outliers, or irrelevant knowledge from both domains. Normally, low-rank reconstruction \cite{shao2014generalized,ding2014latent,jhuo2012robust} and common low-rank coding \cite{ding2015deep,shao2014generalized,ding2018deeptransfer} are two kinds of popular strategies in knowledge transfer. 

Third, robust knowledge fusion is different from knowledge transfer in the sense that it concentrates on common knowledge across multiple domains for new learning tasks. The key is to recover the consistent knowledge by exploiting low-rank modeling across different domains. There are various fusion schemes, e.g., early fusion, late fusion \cite{xia2014robust}, decision fusion \cite{tao2017ensemble} and transferable fusion \cite{ding2018generative}, which essentially place low-rank constraints on feature representations, dicriminant models, and learning outcomes.

In this work, we provide a comprehensive review on robust representation learning via low-rank modeling from a knowledge flow perspective. To our best knowledge, this is the first work to discuss low-rank modeling in such a unified knowledge flow perspective. Specifically, a unified modeling will be introduced to cover most existing models, followed by detailed introduction of representative methods in each line. In addition, extensive discussions and remarks will be provided to further inspire the readers.

\section{A Unified Model}

Given a data set $\mathcal{X} = \{X, Y\}$, where data $X$ or label $Y$ would have noise, incompleteness, and $X$ could contain $v$ subsets with different distributions $\{X_1, \cdots, X_v\}$, we aim to discover robust knowledge from the data by dealing with these challenges based on the specific tasks. The core idea of those existing robust representation learning methods is to explore low-rank modeling to recover, transfer, and fuse knowledge (See Figure \ref{fig1}). We will embody them in two categories: single data domain \textit{w/o} label information; multiple domains \textit{w/o} label information. For clarity, Table \ref{notation} lists the frequently used notations if there are no specific definitions in the main manuscript. 

For \textbf{the first category}, the goal is to recover robust knowledge from single-domain data with low-rank constraints, that is, $X$ contains only data samples from a single domain with similar distribution (i.e., $v=1$). Since the real-world data always suffers from noise or incompleteness, robust knowledge recovery attempts to detect noise and compensate the missing data issues, which can be roughly formulated as the following unified framework:
\renewcommand{\arraystretch}{1.15}
\begin{equation}\label{o1}
\begin{array}{c}
\min\limits_{\phi_{1/2}(\cdot), Z, E} \mathrm{rank}(Z)+\lambda\|E\|_p+\gamma \mathcal{R}\Big(\phi_{1/2}(X), Z, Y\Big)\\
\mathrm{s.t.}~~ \phi_1(X) = \phi_2(X)Z+E,
\end{array}
\end{equation}
where $\phi_{1/2}(\cdot)$ are general mapping functions, e.g., identity function, linear projections, or non-linear functions. $\mathcal{R}(\phi_{1/2}(X), Z, Y)$ denotes a regularizer on the data itself or low-rank representation, or label information, e.g., a task-driven cross-entropy loss, or a graph regularizer \cite{li2016learning}. In Section \ref{Sec3}, we will discuss it more specifically in four lines including low-rank representation learning in original space or transformed space, missing data completion, and graph reconstruction. 

For \textbf{the second category}, we need to either fuse the knowledge or transfer knowledge or both. When data are sampled from different distributions, we can obtain two general problems, one is multi-view learning and the other is transfer learning. For multi-view learning, the goal is to fuse knowledge across multiple views to boost the learning tasks, where data samples are normally correspondent at the sample level. In this sense, we can extend the low-rank constraint in Eq.~\eqref{o1} for each view as $\phi(X_i) = \phi(X_i)Z_i+E_i$, and then either separately fuse knowledge or jointly fuse knowledge by building different strategies over all $Z_i$. For transfer learning, the goal is to adapt well-labeled source knowledge to facilitate the target learning, where data samples across different domains are normally correspondent at the category level. Thus, we can roughly extend the low-rank representation in Eq.~\eqref{o1} as $\phi_1(X_{1/2}) = \phi_2(X_{2/1})Z+E$ to align two domains under low-rank constraint to transfer the intrinsic locality knowledge. Interestingly, zero-shot learning is a special case for transfer learning and multi-view learning, and thus it can be referred to as transferable fusion. In Sections \ref{Sec4} and \ref{Sec5}, we will present knowledge transfer and fusion more specifically.

\begin{table}
\begin{center}\vspace{2mm}
\renewcommand{\arraystretch}{1.35}
\begin{tabular}{cl}
  \Xhline{1pt}
  Notation & Description\\
  \hline
   	$\|\cdot\|_\mathrm{F}$ & Frobenius norm of a matrix\\

  	$\mathrm{rank}(\cdot)$ & Rank operator of a matrix\\

    $\|\cdot\|_{p}$ & The $l_{p}$-norm of a matrix\\

    \hline
    ${\phi}(\cdot), {\phi}_{1/2}(\cdot)$ & General mapping functions\\
    $P/W$ & Linear projection/rotation \\
    
    $F_{1/2}(\cdot)$ & Non-linear functions\\

   	$Z_i$ & The new representation for the $i$-th domain\\

    $Z$ & The Low-rank Representation  \\
    
    $E$ & The Residual \\
    
    $E_i$ & The Residual for the the $i$-th domain\\

    $\lambda, \gamma$ & Balance parameters\\

  \Xhline{1pt}
\end{tabular}
\end{center}\vspace{-3mm}\caption{Notations and Descriptions.}\vspace{-3mm}\label{notation}
\end{table}

\section{Robust Knowledge Recovery}\label{Sec3}

In the learning tasks, we always confront that the real-world data contains noise or incomplete values due to the collected devices and incorrect annotations by human. All these issues hinder us to recover the structure of data, which intrinsically has a very low rankness. Thus, it is essential to recover its low-rank structure that will benefit the learning tasks. 

\subsection{Robust Representation Learning}

In the recent research, robust representation learning has been widely explored when the data contains noise or outliers. Among them, Robust Principal Component Analysis (RPCA) \cite{candes2011robust} is a well-known algorithm, which assumes the data is drawn from one single subspace. Specifically, RPCA decomposes $X$ with $X = Z+E$, where $Z$ is constrained with rank operator and $E$ is the residual with sparse constraint. In this sense, a clean basis could be obtained. Since the real-world data is generally lying in multiple subspaces, Low-Rank Representation (LRR) \cite{liu2013robust} is developed to uncover the global class structures within the data and meanwhile detect sparse noises or outliers. To be specific, LRR learns the low-rank representation $Z$ in the following equation: $X = XZ+E$. RPCA has been successfully applied to background modeling, while LRR achieves impressive performance on subspace clustering. When data is more complex or incomplete, dual low-rank structures are explored to handle this issue. Normally, $X$ will be decomposed to two low-rank components \cite{liu2011latent,panagakis2016robust}. Specifically, latent low-rank representation \cite{liu2011latent} is designed as $X = XZ+WX+E\rightarrow (\mathrm{I}-W)X = XZ+E$, where $Z$ is to recover the structure at sample level and $W$ is to capture the structure at feature level. In this viewpoint, $W$ is considered as a rotation at the feature level. To further preserve the intrinsic structure of the original data, a graph regularizer would be explored to guide the learning of $Z$ and $W$ \cite{yin2015dual}. 

\subsection{Robust Knowledge Embedding}

The previous robust representation recoveries knowledge in the original data space by digging out the noises or outliers. A more effective way to remove the redundancy within data is to jointly learn a transformation or a series of functions. There are considerable research efforts to explore the robust knowledge discovery jointly with embedding techniques, e.g., subspace learning, linear regression, and auto-encoder as well as their variants. 

First of all, robust subspace learning attempts to jointly seek an effective low-dimensional projection $P$ by recovering the intrinsic structure of the noisy data with low-rank constraint \cite{li2016learning}. Among them, we previously explored a supervised learning regularizer $\mathcal{R}(P^\top XZ)$ based on the idea of Fisher criterion to guide the new representation learning, which still exploited the low-rank representation learning in the original space. Moreover, we explored the low-rank representation under the low-dimensional space, i.e., $P^\top X = P^\top XZ+E$, and a graph regularizer $\mathcal{R}(P^\top X)$ is adopted to preserve more discirminative information. Furthermore, we proposed a dual low-rank decomposition for cross-modal data analysis by considering two kinds of manifolds existed within the original data \cite{ding2016robustaaai}. 

Secondly, robust regression aims to trigger a disriminative regression model based on the low-rank representation of the input data, that is, $\|X-Z\|_\mathrm{p}+\lambda\mathrm{rank}(Z)$. Simultaneously, a robust linear regression model $Y = P^\top Z$ is built to recover more discriminative knowledge \cite{huang2016robust}, where $Y$ is the label information and $P$ is the regression model. A tensor-based regression is also proposed by seeking a low-rank tensor transformation \cite{rabusseau2016low}. Other kinds of disriminative pre-defined matrices are also adopted as masks for the specific variables in the optimization formula. To sum up, the key idea of this strategy is to enhance the robustness of regression models by exploring the low-rank representation. 

Thirdly, we previously proposed a robust denoising auto-encoder to reconstruct the clean data with rank constraint \cite{ding2016deep}. The idea behind is that a clean basis will be recovered with low-rank constraint through a series of encoders and decoders with the corrupted data as input. Suppose $F_1(\cdot)$ and $F_2(\cdot)$ are the encoder and decoder, a basic version of robust denoising auto-encoder could be $\|Z - F_2(F_1(X))\|_\mathrm{F}^2$. To integrate more intrinsic structure or discrimiantive information, we design a graph regularizer to guide the encoder learning, i.e., $\mathcal{R}(F_1(X))$. Following this, we explore a robust metric learning by connecting with marginalized desnoising auto-encoder, which aims to seek one linear feature transformation in a marginalized fashion to replace the encoder and decoder of original auto-encoder. To be specific, a robust metric $W \in \mathbb{S}_+^d$ is sought by pursuing low-rank reconstruction as: $\|Z - WX\|_\mathrm{F}^2 = \|Z- PP^\top X\|_\mathrm{F}^2$. 

\subsection{Robust Knowledge Completion}

Since information missing is a very common challenge for data analysis, which is caused by the partial observation on the data. When data has missing values and incompleteness, we need to recover the missing values by borrowing knowledge from the existing values. Specifically, it is essential to capture the structure of the data, and then it will be more likely to recover those missing values. Normally, we will have the following formulation:
\renewcommand{\arraystretch}{1.1}
\begin{equation}\label{miss}
\begin{array}{c}
\min\limits_{Z}\|\mathcal{P}_\Omega (X-Z)\|_\mathrm{F}^2+\lambda\mathrm{rank}(Z),
\end{array}
\end{equation}
where we define the projection $\mathcal{P}_\Omega (X)$ to be the matrix with the observed elements of $X$ preserved, and the missing entries replaced \cite{hastie2015matrix}. 

Another very common case is the incomplete annotation of data. For instance, the labels for each instance are not always well annotated in multi-label learning. That is, to manage the knowledge contained within the data, it is essential to recover the missing labels for each instance by exploring the label relationship across different instances. Normally, the knowledge completion would be constrained on feature-label mapping \cite{jing2015semi} as,
\begin{equation*}\label{miss1}
\begin{array}{c}
\min\limits_{P}\|P^\top X -Y\|_\mathrm{F}^2+\lambda\mathrm{rank}(P^\top X),
\end{array}
\end{equation*}
or directly on multi-label matrix \cite{zhao2015semi} as, 
\begin{equation*}\label{miss1}
\begin{array}{c}
\min\limits_{P, \hat{Y}}\|P^\top X -\hat{Y}\|_\mathrm{F}^2+\lambda\mathrm{rank}(\hat{Y})+\gamma\|\hat{Y}-Y\|_\mathrm{F}^2,
\end{array}
\end{equation*}
where those kinds of methods could be treated as the extensions of data completion and robust regression.

\vspace{2mm}\noindent\textbf{Remarks.} The four lines we discussed previously mainly aim to seek low-rank representations by removing the noise or outliers, reducing the redundant features, and completing the missing elements. They assume that data has a sample-wise low-rank or feature-wise low-rank structure. Specifically, robust representation and completion are two fundamental spirits of robust knowledge discovery. Then, other techniques like subspace learning and linear regression are integrated to capture the robust knowledge in an inductive fashion for convenient knowledge extension. The low-rank knowledge recovery can also be applied to scalable robust graph representation to account for intra- and inter-graph structures in a large-scale graph through divide-and-conquer.

\section{Robust Knowledge Transfer}\label{Sec4}

In learning problems, we are faced with limited training data, leading to poor performance and limited generalization capability. One feasible solution is to reuse knowledge from a relevant field, so called ``domain''. Transfer learning is such a method that uses knowledge from a source domain to assist learning tasks in the target domain, where a good alignment between source and target domains is required. Normally, transfer learning involves two domains $X_s$ and $X_t$, where $X_s$ is well-labeled with $Y_s$. 

\subsection{Robust Transferable Embedding}
In our recent research, we explore a few ways of using low-rank representation together with subspace learning for domain alignment called ``Low-rank Transfer Subspace Learning (LSTL)'' \cite{shao2014generalized}. Assume the source domain data is $X_s$ and target domain data is $X_t$, the proposed LSTL aims to find a subspace $P$ and a low-rank alignment matrix $Z$, such that the following mapping is valid:
\renewcommand{\arraystretch}{1.5}
\begin{equation}
\begin{array}{c}
\min\limits_{Z,E,P} \mathrm{rank}(Z) + \lambda \|E\|_{2,1} + \gamma f(P, X_s, X_t),\\
\mathrm{s.t.}~~P^\top X_s = P^\top X_tZ+E. 
\label{eq: LSTL}
\end{array}
\end{equation}

The model above not only allows knowledge transfer but also discovers a suitable subspace at the same time. The term $E$ aims to account for the ``outliers'' that cannot be well incorporated into the knowledge transfer, which leads to a robust and explainable learning process. The optimization problem here is not jointly convex over all unknown variables. Thus, ADMM method is usually applied to update each unknown variable at a time by fixing others until all converged. 

It should be noted that the model in Eq.~\eqref{eq: LSTL} is generic in the sense that it encapsulates many existing subspace learning methods. By adjusting the learning function $f(P,X_S, X_t)$, it will adapt to supervised or unsupervised subspace learning, and address the source domain with or without labels. In addition, it deserves further exploration of using kernel or tensor structure which may explain the non-linear or multilinear data well. This method has been thoroughly evaluated on various tasks including domain adaptation, heterogeneous face recognition, and visual kinship verification \cite{shao2014generalized}.

There are three typical model variations similar to LSTL. The first model simplifies LSTL and degenerates to learning a rotation function only instead of enforcing subspace projection $P$ on both sides of Eq.~\eqref{eq: LSTL}. Specifically, they use the constraint: $WX_s = X_tZ+E$ where $W$ denotes the rotation matrix \cite{jhuo2012robust}, which aims to rotate the source domain and align with target domain under low-rank reconstruction. Second, a more challenging yet valuable problem recently draws our attention that the target domain is completely missing in training period, in which conventional transfer learning models may fail. To compensate for the missing target, we will borrow the knowledge from a complete source-target transfer learning scenario as a complement for the original LSTL \cite{ding2014latent}. 
Finally, to further explore robust embedding in unsupervised scenario, we designed a new transfer subspace clustering method for motion segmentation with a weighted rank constraint \cite{wang2018low}. Different from original low-rank constraint on the coefficients, we convert traditional graph regularizer to a weighted rank optimization to intuitively seek more intrinsic structure of the data.

\vspace{2mm}\noindent\textbf{Remarks.} The approaches discussed above refer to a common subspace and thus work fairly well in various vision tasks, e.g., visual domain adaptation. It may serve as a feature extraction module at the same time, if the input is the raw feature, e.g., pixel values of a face image. But its performance can be significantly improved if using recently developed deep features such as VGG-Net and ResNet. We realized that some recent methods integrate knowledge transfer into deep learning, and formulate an end-to-end learning paradigm for better performance \cite{ding2018deeptransfer}. We believe this learning strategy performs well given sufficient target domain data, and may require labeled target similar to fine-tuning. In the case of limited target data given and large domain divergence, the end-to-end training may not work well. Differently, our model is able to handle both labeled and unlabeled source/target data (depending on the subspace learning function) and it only focuses on representation learning. Second, it should point out the computational cost in Eq.~\eqref{eq: LSTL} and its variants is usually heavy due to matrix multiplications and SVD decomposition, and methods like proximal gradient decent or random projection may help.

\subsection{Robust Transferable Coding}

Beyond robust transferable embedding, robust transferable coding is also a very appealing strategy to seek domain-invariant representation across different domains. The very pioneering work is robust sparse transfer coding, whose idea is to seek domain-invariant sparse coding for both domains and the objective function can be formulated as $\|X-DZ\|_\mathrm{F}^2+\gamma\|Z\|_1+\alpha \mathcal{R}(Z)$ \cite{long2013transfer}. Following this, Al-Shedivat et al. extended the transferable sparse coding and proposed a task-driven manner by incorporating a SVM-based term \cite{al2014supervised}. 

However, sparse coding cannot capture the global structure of the data. Thus, exploiting low-rank coding in knowledge transfer is an effective strategy. Following this, we proposed a low-rank transfer coding scheme by seeking better class-wise structure across two domains. Inspired by layer-wise training philosophy, more effective and domain-invariant representation would be learned through a hierarchical structure. Specifically, we explore a layer-wise strategy to build multi-layer representation to implement knowledge transfer. A marginalized auto-encoder $W$ is jointly learned to transform the data into the new low-rank coding as $WX = WX_sZ+E$ where $Z$ is the layer-wise low-rank coding \cite{ding2015deep}. Following this, Shao et al. first organized the source-target pairs using decision tree and then sought low-rank coding for both domains \cite{shao2016spectral}.

In knowledge transfer, it may occur that the source data is unlabeled while limited target data is labeled, which is referred as ``self-taught learning''. To address this problem, we explore self-taught low-rank (S-Low) coding to effectively utilize the rich low-level pattern information abstracted from the auxiliary domain and characterize the high-level structural information in the target domain \cite{li2017self}. 
Since many types of visual data have been proven to contain subspace structures, a low-rank constraint is deployed into the coding objective to better characterize the structure of the given target set.

\subsection{Robust Transferable Networks}

Recent research efforts have revealed that deep structure learning is able to generate more domain-invariant features for knowledge transfer with promising performance on existing cross-domain benchmarks. Deep transfer learning aims to unify knowledge transfer and deep feature extraction into one training procedure, and they aim to capture a general domain-invariant but task-discriminative representation in the latent feature space by matching the source and target domain data distributions, e.g., discrepancy loss or adversarial loss.

Along this line, we extended our stacked low-rank coding scheme to an end-to-end transferable network by seeking multi-layer low-rank coding through matrix factorization at the top layer \cite{ding2018deeptransfer}. Specifically, we adapt previous low-rank coding into multiple layers as $\|X-D_1D_2\cdots{D_k}Z_k\|_\mathrm{F}^2+\gamma\mathrm{rank}(Z_k)$, where the learned representation in each layer is able to progressively capture the domain invariant features, and provide better performance in general compared to the shallow approaches.

Most recently, Chen et al. discussed the relationship between transferability and discriminability through batch spectral penalization \cite{chen2019transferability}, which cast a light for rank minimization into mini-batch strategy. Their work indicates that the eigenvectors with the largest singular values will dominate the feature transferability. As a consequence, the transferability is enhanced at the expense of over penalization of other eigenvectors that embody rich structures crucial for discriminability. 

\vspace{2mm}\noindent\textbf{Remarks.} When applying rank constraint to neural networks learning, heavy computation cost would be a big concern. Divide-and-Conquer tends to be a promising strategy. Another way is batch-based rank constraint. Since graph neural networks are popular in capturing intrinsic structure of the data, how to build a more robust graph is very essential, especially under mini-batch strategy. 

\section{Robust Knowledge Fusion}\label{Sec5}

A great deal of attention, as of recent, has been focused on multi-source or multi-view learning, as these data is of abundance in reality. When data is collected from different viewpoints, or captured by different sensors, the data of the same class will render very differently in terms of appearance, or mid-level representation. Thus, conventional way of data fusion by simply concatenating multi-view or multi-source data will not work well, let alone the heterogeneous knowledge that are compatible with each other by nature. Finding consistent yet discriminant representations for these multi-view data motives an interesting yet challenging problem called knowledge fusion. According to the stage in the knowledge fusion, we may briefly categorize into: early fusion, decision fusion, and late fusion, corresponding to different AI problems including clustering, outlier detection, classification, etc. Next, we will discuss how to integrate low-rank modeling with knowledge fusion.

In knowledge fusion, the misalignment in multi-view or heterogeneous features would significantly affect the performance in learning tasks. The first feasible attempt would be finding a common representation. Canonical correlation analysis (CCA) and related coupled projection methods have been dominant in this field, and their variations including kernel or tensor extensions achieve promising performance in complex data. Here we will illustrate robust knowledge fusion through rank minimization. Two straightforward ways to explore the previous robust knowledge representation are early fusion and late fusion. Early fusion attempts to discovery the knowledge after we concatenate all-view data together. Late fusion aims to fuse the multi-view knowledge after some pre-processing steps, as the name suggests. Different applications would first learn multiple outputs based on various tasks, and then a common low-rank output is obtained through low-rank and sparse decomposition by digging out the view-specific sparse residual per view \cite{xia2014robust}.

\subsection{Robust Decision Fusion}

Another main-stream fusion strategy is decision fusion, which has been widely explored in robust knowledge fusion. This fusion strategy jointly explores the representation learning and multi-view fusion in a unified framework. Thus, we have more flexibility to fuse knowledge across different views, but will suffer from higher complexity during the optimization. Similar to the single-view data analysis, multi-view learning aims to fuse the knowledge across different views by jointly seeking common low-rank representation, robust embedding and completion. Thus, multi-view knowledge fusion is an extension of single-view knowledge discovery. The key is how to deliver the fusion strategy to capture the consistent knowledge across multiple views. 

First of all, there are mainly three strategies to recover a consistent representation, which can be formulated by the following unified learning model:
\renewcommand{\arraystretch}{1.3}
\begin{equation}\label{fusion1}
\begin{array}{c}
\min\limits_{Z_i, E_i} \sum\limits_{i=1}^v\mathcal{F}(Z_i) + \lambda \|E_i\|_{p},\\
\mathrm{s.t.} ~~X_i = X_iZ_i+E_i,
\end{array}
\end{equation}
where $\sum_{i=1}^v\mathcal{F}(Z_i)$ is to deliver the knowledge fusion for $v$ views or domains. The first strategy is to further explore low-rank decomposition to seek a common low-rank representation and minimize the residuals of each view-specific representation and common one \cite{li2018latent}, i.e., 
\begin{equation*}\label{fusion11}
\begin{array}{c}
\sum\limits_{i=1}^v\mathcal{F}(Z_i) = \mathrm{rank}(Z)+\gamma\sum\limits_{i=1}^v\|Z_i-Z\|_p.
\end{array}
\end{equation*}

The second strategy is to explore matrix factorization to further factorize the view-specific representation into one common part and another view-specific one \cite{wang2018multiview}, i.e., 
\begin{equation*}\label{fusion12}
\begin{array}{c}
\sum\limits_{i=1}^v\mathcal{F}(Z_i) =\sum\limits_{i=1}^v\Big(\|Z_i-U_iV\|_\mathrm{F}^2+\dfrac{1}{2}\|U_i\|_\mathrm{F}^2\Big)+\dfrac{1}{2}\|V\|_\mathrm{F}^2,
\end{array}
\end{equation*}
where $U_iV$ is a low-rank approximation of $Z_i$ and $V$ is the common knowledge across different views. Following this, we proposed a multi-view clustering algorithm with ensemble strategy, adopting multiple co-association matrices as the input to seek a low-rank common representation \cite{tao2017ensemble}. The third is to explore some regualarizers to merge each low-rank view-specific representation as a whole into the optimization \cite{cheng2011multi}, e.g., 
\begin{equation*}\label{fusion12}
\begin{array}{c}
\sum\limits_{i=1}^v\mathcal{F}(Z_i) = \sum\limits_{i=1}^v\mathrm{rank}(Z_i)+\gamma\mathcal{M}(Z),
\end{array}
\end{equation*}
where $\mathcal{M}(Z)$ is the knowledge fusion term over multiple $Z_i$.

Second, embedding techniques are always exploited in knowledge fusion, since available multi-view data are always lying in different feature spaces. In other words, multiple view-specific transformations are able to project the data into a common space, where knowledge could be fused through low-rank representation simultaneously. Along this strategy, we proposed a low-rank collective subspace by capturing feature-level and sample-level consistent knowledge across multiple views \cite{ding2018robustmultiview}. 

Third, data missing issue is much more common within multi-view data than single-view data. In this sense, there are two tasks going together, i.e., knowledge fusion and completion. Along this line, Zhang et al.~explored CCA to fuse the cross-view knowledge using low-rank matrix to recover the missing knowledge \cite{zhang2018multi}. While Liu et al. directly explored a multi-view regression model to recover the consistent knowledge for multi-view multi-label image classification \cite{liu2015low}.

\subsection{Robust Transferable Fusion}

In more complex learning problems, we would consider knowledge transfer and fusion simultaneously. Here, we discuss low-rank modeling applied to three highly related learning tasks, including zero-shot learning, multi-source domain adaptation and domain generalization. 

Zero-shot learning (ZSL), as a special case of multi-view learning and transfer learning, is inspired by the learning mechanism of human brain \cite{liu2018zero,ding2018generative}. The goal is to classify new categories which are unobserved during the training process. For instance, one is able to predict a new species of animal after being informed what it looks like and how it is different from or similar to other known animals. Generally, there will be two views in ZSL, i.e., visual features and semantic features that are highly coupled. Due to the uniqueness of ZSL, two aspects need to be considered, i.e., knowledge fusion across two views and knowledge transfer from seen data to unseen one. Recently, low-rank constraint has been explored in ZSL to equip a low-rank embedding with highlighting the shared semantics across different seen categories \cite{ding2018generative,liu2018zero}. The idea of low-rank constraint is to group such similar visual features underlying the embedding space. In this way, discriminative and descriptive features from seen categories could be adapted to the unseen ones. 

Different from zero-shot learning, multi-source adaptation and domain generalization include the data from multiple domains but without pair-wise correlation. In this sense, the fusion strategy would target at coupling the category-wise knowledge across multiple domains. Previous robust domain alignment is still valid for each pair of source and target \cite{ding2018incomplete} or a shared dictionary \cite{ding2018robustmultiview}, which assumes to carry the shared knowledge to target domain. Beyond that, discriminative cross-source regularizers are designed to fuse the knowledge from multiple sources \cite{ding2018incomplete,ding2018robustmultiview}, since the cross-source samples are lying in a generic space or similar feature spaces. 
Furthermore, previous robust transferable networks can also be explored to build an end-to-end trainable architecture, where we exploited low-rank constraint in the hidden layers to guide the domain-specific and domain-shared networks learning, which can assist better transfer knowledge from multiple sources to boost the learning problem in unseen target domains \cite{ding2017deep_dg}.

\vspace{2mm}\noindent\textbf{Remarks.} While previous sections mainly discuss how to robustly fuse the knowledge in multi-view/multi-domain data analysis, we emphasize on \textit{decision fusion} and \textit{transferable fusion}, that tend to be popular in real-world applications. When deploying deep neural networks to extract the features of multi-view/multi-domain data, robust knowledge fusion still plays a major role in the entire feature learning pipeline.

\section{Conclusion \& Future Direction}

In this paper, we presented a comprehensive survey on robust visual representation learning via low-rank modeling, where we mainly covered single-domain knowledge and multi-domain knowledge discovery. We identified the shared and distinct terms across them, and presented detailed discussion including our recently proposed algorithms for robust knowledge recovery, robust knowledge transfer and robust knowledge fusion. This would benefit the AI community in both industry and academia from literature review to future directions. Despite the recent advances, some potential future research directions on robust knowledge fusion and transfer are listed as follows:

{First}, large-scale multi-view image retrieval needs many image pairs across views to learn correspondence. But both the probe and reference images are not under control in terms of both quality and quantity. For example, in forensic face recognition, we have a single sketch face as reference to retrieve RGB faces from surveillance cameras, and enrolled face images from police department. The sketch needs to be converted to common feature first and then compared against RGB faces. The single sketch and many other RGB faces from different persons, with varied quality and numbers pose an extreme unbalanced learning. 

{Second}, how to adapt the knowledge from existing large-scale public datasets to new domains or problems where training samples are few? This is extremely critical for problems that need knowledge extrapolation. This is essentially a ``compound'' of few-shot learning and domain adaptation. We still take face recognition as an example, where we intend to extend the well-trained face recognition algorithms for day time to night light under poor illuminations. We may only given few images per person in night time, which accounts for the extremely incomplete multi-view data.

\newpage
\small
\bibliographystyle{named}
\bibliography{ijcai20}

\begin{thebibliography}{}

\bibitem[\protect\citeauthoryear{Al-Shedivat \bgroup \em et al.\egroup
  }{2014}]{al2014supervised}
Maruan Al-Shedivat, Jim Jing-Yan Wang, Majed Alzahrani, Jianhua~Z Huang, and
  Xin Gao.
\newblock Supervised transfer sparse coding.
\newblock In {\em AAAI}, 2014.

\bibitem[\protect\citeauthoryear{Cand{\`e}s \bgroup \em et al.\egroup
  }{2011}]{candes2011robust}
Emmanuel~J Cand{\`e}s, Xiaodong Li, Yi~Ma, and John Wright.
\newblock Robust principal component analysis?
\newblock {\em JACM}, 58(3):11, 2011.

\bibitem[\protect\citeauthoryear{Chen \bgroup \em et al.\egroup
  }{2019}]{chen2019transferability}
Xinyang Chen, Sinan Wang, Mingsheng Long, and Jianmin Wang.
\newblock Transferability vs. discriminability: Batch spectral penalization for
  adversarial domain adaptation.
\newblock In {\em ICML}, pages 1081--1090, 2019.

\bibitem[\protect\citeauthoryear{Cheng \bgroup \em et al.\egroup
  }{2011}]{cheng2011multi}
Bin Cheng, Guangcan Liu, Jingdong Wang, Zhongyang Huang, and Shuicheng Yan.
\newblock Multi-task low-rank affinity pursuit for image segmentation.
\newblock In {\em ICCV}, pages 2439--2446, 2011.

\bibitem[\protect\citeauthoryear{Ding and Fu}{2016}]{ding2016robustaaai}
Zhengming Ding and Yun Fu.
\newblock Robust multi-view subspace learning through dual low-rank
  decompositions.
\newblock In {\em AAAI}, 2016.

\bibitem[\protect\citeauthoryear{Ding and Fu}{2017}]{ding2017deep_dg}
Zhengming Ding and Yun Fu.
\newblock Deep domain generalization with structured low-rank constraint.
\newblock {\em IEEE TIP}, 27(1):304--313, 2017.

\bibitem[\protect\citeauthoryear{Ding and Fu}{2018a}]{ding2018deeptransfer}
Zhengming Ding and Yun Fu.
\newblock Deep transfer low-rank coding for cross-domain learning.
\newblock {\em IEEE TNNLS}, 2018.

\bibitem[\protect\citeauthoryear{Ding and Fu}{2018b}]{ding2018robustmultiview}
Zhengming Ding and Yun Fu.
\newblock Robust multiview data analysis through collective low-rank subspace.
\newblock {\em IEEE TNNLS}, 29(5):1986--1997, 2018.

\bibitem[\protect\citeauthoryear{Ding \bgroup \em et al.\egroup
  }{2014}]{ding2014latent}
Zhengming Ding, Ming Shao, and Yun Fu.
\newblock Latent low-rank transfer subspace learning for missing modality
  recognition.
\newblock In {\em AAAI}, pages 1192--1198, 2014.

\bibitem[\protect\citeauthoryear{Ding \bgroup \em et al.\egroup
  }{2015}]{ding2015deep}
Zhengming Ding, Ming Shao, and Yun Fu.
\newblock Deep low-rank coding for transfer learning.
\newblock In {\em IJCAI}, 2015.

\bibitem[\protect\citeauthoryear{Ding \bgroup \em et al.\egroup
  }{2016}]{ding2016deep}
Zhengming Ding, Ming Shao, and Yun Fu.
\newblock Deep robust encoder through locality preserving low-rank dictionary.
\newblock In {\em ECCV}, pages 567--582, 2016.

\bibitem[\protect\citeauthoryear{Ding \bgroup \em et al.\egroup
  }{2018a}]{ding2018generative}
Zhengming Ding, Ming Shao, and Yun Fu.
\newblock Generative zero-shot learning via low-rank embedded semantic
  dictionary.
\newblock {\em IEEE TPAMI}, 2018.

\bibitem[\protect\citeauthoryear{Ding \bgroup \em et al.\egroup
  }{2018b}]{ding2018incomplete}
Zhengming Ding, Ming Shao, and Yun Fu.
\newblock Incomplete multisource transfer learning.
\newblock {\em IEEE TNNLS}, 2018.

\bibitem[\protect\citeauthoryear{Hastie \bgroup \em et al.\egroup
  }{2015}]{hastie2015matrix}
Trevor Hastie, Rahul Mazumder, Jason~D Lee, and Reza Zadeh.
\newblock Matrix completion and low-rank svd via fast alternating least
  squares.
\newblock {\em JMLR}, 16(1):3367--3402, 2015.

\bibitem[\protect\citeauthoryear{Huang \bgroup \em et al.\egroup
  }{2016}]{huang2016robust}
Dong Huang, Ricardo Cabral, and Fernando De~la Torre.
\newblock Robust regression.
\newblock {\em IEEE TPAMI}, 38(2):363--375, 2016.

\bibitem[\protect\citeauthoryear{Jhuo \bgroup \em et al.\egroup
  }{2012}]{jhuo2012robust}
I-Hong Jhuo, Dong Liu, DT~Lee, and Shih-Fu Chang.
\newblock Robust visual domain adaptation with low-rank reconstruction.
\newblock In {\em CVPR}, pages 2168--2175, 2012.

\bibitem[\protect\citeauthoryear{Jing \bgroup \em et al.\egroup
  }{2015}]{jing2015semi}
Liping Jing, Liu Yang, Jian Yu, and Michael~K Ng.
\newblock Semi-supervised low-rank mapping learning for multi-label
  classification.
\newblock In {\em CVPR}, pages 1483--1491, 2015.

\bibitem[\protect\citeauthoryear{Li and Fu}{2016}]{li2016learning}
Sheng Li and Yun Fu.
\newblock Learning robust and discriminative subspace with low-rank
  constraints.
\newblock {\em IEEE TNNLS}, 27(11):2160--2173, 2016.

\bibitem[\protect\citeauthoryear{Li \bgroup \em et al.\egroup
  }{2017}]{li2017self}
Sheng Li, Kang Li, and Yun Fu.
\newblock Self-taught low-rank coding for visual learning.
\newblock {\em IEEE TNNLS}, 29(3):645--656, 2017.

\bibitem[\protect\citeauthoryear{Li \bgroup \em et al.\egroup
  }{2018}]{li2018latent}
Kai Li, Sheng Li, Zhengming Ding, Weidong Zhang, and Yun Fu.
\newblock Latent discriminant subspace representations for multi-view outlier
  detection.
\newblock In {\em AAAI}, 2018.

\bibitem[\protect\citeauthoryear{Liu and Yan}{2011}]{liu2011latent}
Guangcan Liu and Shuicheng Yan.
\newblock Latent low-rank representation for subspace segmentation and feature
  extraction.
\newblock In {\em ICCV}, pages 1615--1622, 2011.

\bibitem[\protect\citeauthoryear{Liu \bgroup \em et al.\egroup
  }{2013}]{liu2013robust}
Guangcan Liu, Zhouchen Lin, Shuicheng Yan, Ju~Sun, Yong Yu, and Yi~Ma.
\newblock Robust recovery of subspace structures by low-rank representation.
\newblock {\em IEEE TPAMI}, 35(1):171--184, 2013.

\bibitem[\protect\citeauthoryear{Liu \bgroup \em et al.\egroup
  }{2015}]{liu2015low}
Meng Liu, Yong Luo, Dacheng Tao, Chao Xu, and Yonggang Wen.
\newblock Low-rank multi-view learning in matrix completion for multi-label
  image classification.
\newblock In {\em AAAI}, 2015.

\bibitem[\protect\citeauthoryear{Long \bgroup \em et al.\egroup
  }{2013}]{long2013transfer}
Mingsheng Long, Guiguang Ding, Jianmin Wang, Jiaguang Sun, Yuchen Guo, and
  Philip~S Yu.
\newblock Transfer sparse coding for robust image representation.
\newblock In {\em CVPR}, pages 407--414, 2013.

\bibitem[\protect\citeauthoryear{Panagakis \bgroup \em et al.\egroup
  }{2016}]{panagakis2016robust}
Yannis Panagakis, Mihalis Nicolaou, Stefanos Zafeiriou, and Maja Pantic.
\newblock Robust correlated and individual component analysis.
\newblock {\em IEEE TPAMI}, 38(8):1665--1678, 2016.

\bibitem[\protect\citeauthoryear{Rabusseau and Kadri}{2016}]{rabusseau2016low}
Guillaume Rabusseau and Hachem Kadri.
\newblock Low-rank regression with tensor responses.
\newblock In {\em NIPS}, pages 1867--1875, 2016.

\bibitem[\protect\citeauthoryear{Shao \bgroup \em et al.\egroup
  }{2014}]{shao2014generalized}
Ming Shao, Dmitry Kit, and Yun Fu.
\newblock Generalized transfer subspace learning through low-rank constraint.
\newblock {\em IJCV}, 109(1-2):74--93, 2014.

\bibitem[\protect\citeauthoryear{Shao \bgroup \em et al.\egroup
  }{2016a}]{shao2016spectral}
Ming Shao, Zhengming Ding, Handong Zhao, and Yun Fu.
\newblock Spectral bisection tree guided deep adaptive exemplar autoencoder for
  unsupervised domain adaptation.
\newblock In {\em AAAI}, 2016.

\bibitem[\protect\citeauthoryear{Shao \bgroup \em et al.\egroup
  }{2016b}]{shao2016scalable}
Ming Shao, Xindong Wu, and Yun Fu.
\newblock Scalable nearest neighbor sparse graph approximation by exploiting
  graph structure.
\newblock {\em IEEE TBD}, 2(4):365--380, 2016.

\bibitem[\protect\citeauthoryear{Tao \bgroup \em et al.\egroup
  }{2017}]{tao2017ensemble}
Zhiqiang Tao, Hongfu Liu, Sheng Li, Zhengming Ding, and Yun Fu.
\newblock From ensemble clustering to multi-view clustering.
\newblock In {\em IJCAI}, pages 2843--2849, 2017.

\bibitem[\protect\citeauthoryear{Wang \bgroup \em et al.\egroup
  }{2018a}]{wang2018low}
Lichen Wang, Zhengming Ding, and Yun Fu.
\newblock Low-rank transfer human motion segmentation.
\newblock {\em IEEE TIP}, 28(2):1023--1034, 2018.

\bibitem[\protect\citeauthoryear{Wang \bgroup \em et al.\egroup
  }{2018b}]{wang2018multiview}
Yang Wang, Lin Wu, Xuemin Lin, and Junbin Gao.
\newblock Multiview spectral clustering via structured low-rank matrix
  factorization.
\newblock {\em IEEE TNNLS}, (99):1--11, 2018.

\bibitem[\protect\citeauthoryear{Xia \bgroup \em et al.\egroup
  }{2014}]{xia2014robust}
Rongkai Xia, Yan Pan, Lei Du, and Jian Yin.
\newblock Robust multi-view spectral clustering via low-rank and sparse
  decomposition.
\newblock In {\em AAAI}, pages 2149--2155, 2014.

\bibitem[\protect\citeauthoryear{Yin \bgroup \em et al.\egroup
  }{2015}]{yin2015dual}
Ming Yin, Junbin Gao, Zhouchen Lin, Qinfeng Shi, and Yi~Guo.
\newblock Dual graph regularized latent low-rank representation for subspace
  clustering.
\newblock {\em IEEE TIP}, 24(12):4918--4933, 2015.

\bibitem[\protect\citeauthoryear{Zhang \bgroup \em et al.\egroup
  }{2018}]{zhang2018multi}
Lei Zhang, Yao Zhao, Zhenfeng Zhu, Dinggang Shen, and Shuiwang Ji.
\newblock Multi-view missing data completion.
\newblock {\em IEEE TKDE}, 30(7):1296--1309, 2018.

\bibitem[\protect\citeauthoryear{Zhao and Guo}{2015}]{zhao2015semi}
Feipeng Zhao and Yuhong Guo.
\newblock Semi-supervised multi-label learning with incomplete labels.
\newblock In {\em IJCAI}, pages 4062--4068, 2015.

\end{thebibliography}

\end{document}